\documentclass{ecai} 

\usepackage{latexsym}
\usepackage{amssymb}
\usepackage{amsmath}
\usepackage{amsthm}
\usepackage{booktabs}
\usepackage{enumitem}
\usepackage{graphicx}
\usepackage{color}

\usepackage{multirow}

\newtheorem{theorem}{Theorem}
\newtheorem{lemma}{Lemma}

\newtheorem{assumption}{Assumption}

\newcommand{\BibTeX}{B\kern-.05em{\sc i\kern-.025em b}\kern-.08em\TeX}

\begin{document}


\begin{frontmatter}


\paperid{6636} 


\title{From Noise to Precision: A Diffusion-Driven Approach to
Zero-Inflated Precipitation Prediction}


\author[A]{\fnms{Wentao}~\snm{Gao}\thanks{Corresponding Author. Email: gaowy014@mymail.unisa.edu.au.}}
\author[A]{\fnms{Jiuyong}~\snm{Li}}
\author[A]{\fnms{Lin}~\snm{Liu}}
\author[A]{\fnms{Thuc Duy}~\snm{Le}}
\author[A]{\fnms{Xiongren}~\snm{Chen}}
\author[A]{\fnms{Xiaojing}~\snm{Du}}
\author[A]{\fnms{Jixue}~\snm{Liu}}
\author[C]{\fnms{Yanchang}~\snm{Zhao}}
\author[B]{\fnms{Yun}~\snm{Chen}}

\address[A]{University of South Australia, Adelaide, SA, Australia}
\address[B]{CSIRO Environment, Canberra, Australia}
\address[C]{CSIRO Data61, Canberra, Australia}


\begin{abstract}
Zero-inflated data pose significant challenges in precipitation forecasting due to the predominance of zeros with sparse non-zero events. To address this, we propose the Zero Inflation Diffusion Framework (ZIDF), which integrates Gaussian perturbation for smoothing zero-inflated distributions, Transformer-based prediction for capturing temporal patterns, and diffusion-based denoising to restore the original data structure. In our experiments, we use observational precipitation data collected from South Australia along with synthetically generated zero-inflated data. Results show that ZIDF demonstrates significant performance improvements over multiple state-of-the-art precipitation forecasting models, achieving up to 56.7\% reduction in MSE and 21.1\% reduction in MAE relative to the baseline Non-stationary Transformer. These findings highlight ZIDF's ability to robustly handle sparse time series data and suggest its potential generalizability to other domains where zero inflation is a key challenge.
\end{abstract}

\end{frontmatter}


\section{Introduction}

\begin{figure}[h]
    \centering
    \includegraphics[width=1\linewidth]{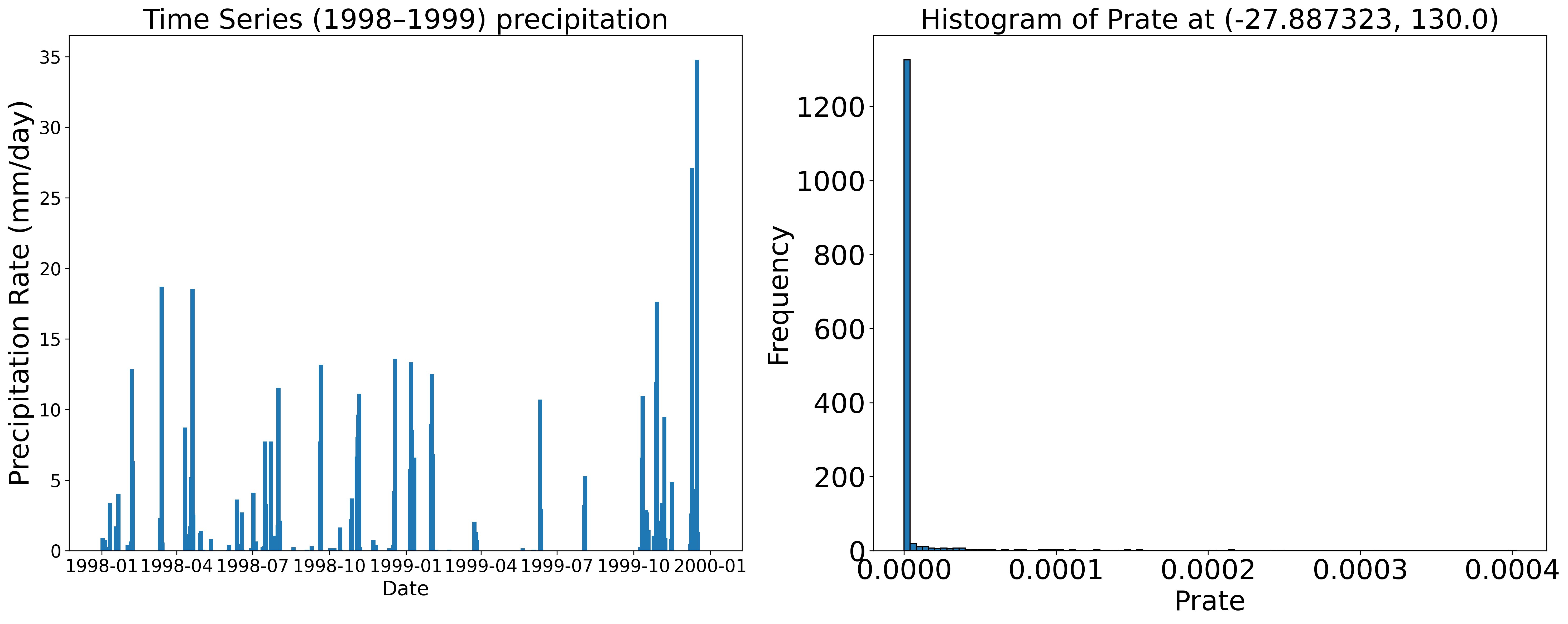}
    \caption{Challenges of Zero-Inflated Drought Data: Visualizing Global Extreme Drought Trends and Sparse Precipitation (prate) Events.}
    \label{fig:precipitation_data}
\end{figure}

Precipitation prediction is crucial for a variety of applications, including meteorology, agriculture, and water resource management, where accurate forecasts inform flood control, drought mitigation, and agricultural planning. However, as shown in Figure~\ref{fig:precipitation_data}, precipitation datasets often exhibit \textit{zero inflation}, meaning that the data contain predominantly zero-valued observations (dry days) punctuated by relatively sparse non-zero precipitation events. This sparse distribution poses a significant challenge: traditional models struggle to identify and accurately predict rare but crucial non-zero events amidst the overwhelming zeros \cite{lambert1992zero,mullahy1986specification, DEMM2021, Feng2023_rare_events_imblanced}. In addition to zero inflation, precipitation data often exhibit \textit{complex temporal dependencies}, influenced by regional climatic patterns, seasonal cycles, and large-scale teleconnections such as the El Niño-Southern Oscillation (ENSO) \cite{Trenberth2000enso}. These complexities demand models capable of capturing intricate temporal structures while also dealing with a large fraction of zero values.

To handle zero inflation, classical statistical approaches like the hurdle \cite{rose2006use} and zero-inflated models \cite{lambert1992zero} have been widely explored \cite{zero_review}. The hurdle model separates the occurrence of zeros and non-zeros into distinct processes but does not differentiate the origin of zeros \cite{kong2020deephurdlenetworkszeroinflated,Feng2021}. In contrast, zero-inflated models explicitly distinguish \textit{structural zeros} (e.g., truly dry days) from \textit{non-structural zeros} (zeros that occasionally appear within the main precipitation distribution) by modeling zeros via a Bernoulli process and non-zeros via a continuous distribution \cite{Sheng2023}. 

In the statistical and actuarial literature, more sophisticated approaches for zero-inflated data have been developed. Tweedie compound Poisson models \cite{dunn2005tweedie,yang2018insurance} 
offer flexibility for data with excess zeros and continuous positive responses, 
and have demonstrated effectiveness in highly imbalanced settings. Gradient tree-boosting combined with Tweedie models \cite{zhou2022tweedie} has demonstrated substantial improvements for extremely unbalanced zero-inflated data. Zero-inflated quantile regression \cite{ling2022statistical} provides an alternative distribution-free approach that focuses on specific quantiles of interest, particularly valuable for extreme event prediction. While these models offer nuanced statistical interpretations of zero generation, they often face difficulties in capturing complex temporal dependencies and large-scale variability inherent in precipitation time series \cite{zero_review}.

Deep learning approaches have shown promise in capturing spatial and temporal patterns in meteorological data \cite{ravuris2021skillful, bi2023accurate}. However, standard deep models like CNNs or GANs typically do not explicitly address zero inflation. They struggle to distinguish between structural and non-structural zeros \cite{DEMM2021}, limiting their capacity to isolate meaningful non-zero events and weakening their predictive power in zero-inflated regimes. More sophisticated frameworks, such as the Deep Extreme Mixture Model (DEMM), integrate hurdle structures with extreme value theory to effectively model zeros, moderate, and extreme events \cite{DEMM2021}. Despite yielding more accurate representations of precipitation distributions, the high parameter complexity of these sophisticated methods significantly increases training and inference costs, reducing their scalability to large datasets.

On the other hand, Transformer-based models have emerged as powerful tools for time series forecasting, leveraging self-attention to handle long-term dependencies efficiently \cite{liu2023nonstationarytransformersexploringstationarity,zhou2021informerefficienttransformerlong,liu2023itransformer}. Transformers, with their simpler structural priors, excel in dealing with a wide range of temporal patterns. Yet, they do not inherently solve the zero-inflated data challenges. As detailed in Section 2.2, directly applying Transformers to zero-inflated data can degrade their performance because the attention mechanism cannot easily discern the nuanced difference between structural zeros and rare non-zero occurrences. This motivates the integration of a different class of models that can effectively reshape data distributions before Transformer-based prediction.

Diffusion models have recently gained attention for their ability to model complex distributions in time series data. For instance, TimeDiff \cite{zhang2023time_diffusion} leverages diffusion processes to learn non-stationary functions and capture intricate temporal structures, while \cite{yoon2024probabilistic} applies diffusion models to probabilistic forecasting, effectively modeling uncertainty. These advances suggest that diffusion models can offer robust distributional learning capabilities. However, their use directly tackling zero inflated time series such as precipitation forecasting remains largely unexplored. By integrating diffusion models into the framework, we can first smooth the zero-inflated distribution, making it more amenable to gradient-based learning methods and improving the downstream performance of models like Transformers.

\textbf{Contributions} In this work, we propose a novel framework to address zero-inflated data challenges in precipitation forecasting by combining Gaussian smoothing \cite{Lindeberg_2024}, Transformer-based prediction, and diffusion-based denoising. Our key contributions include:
\begin{itemize}
    \item \textbf{Gaussian Perturbation for Distribution Smoothing:} We introduce a Gaussian noise injection strategy that transforms the original zero-inflated distribution into a smoother one. This enables stable gradient-based optimization, allowing the predictive component to learn meaningful temporal patterns without being overwhelmed by zeros. \textit{In practice, this improvement leads to more reliable predictions of rare precipitation events, which are crucial for flood warning systems.}
    
    \item \textbf{Integrated Diffusion Model Framework:} We propose a diffusion model-based framework ZIDF (Zero Inflation Diffusion Framework) tailored for zero-inflated time series data. After training a predictive component (e.g., a Non-stationary Transformer) for the smoothed targets, we employ a diffusion-based denoising process. \textit{This two-step approach reduces MSE by up to 56.7\% compared to direct prediction, significantly improving forecast performance for agricultural planning and water resource management.}
    
    \item \textbf{Efficient Denoising Mechanism:} By leveraging the inverse diffusion steps, we effectively remove the injected noise, recovering the original distribution from the model's noisy predictions. \textit{This procedure enhances the detection of sparse but critical non-zero precipitation events, crucial for drought monitoring and extreme weather preparedness.}
\end{itemize}

\section{Preliminaries}

\subsection{Problem Statement}
Our problem is a time series forecasting task with a large number of zero values in the target variable $Y$, which represents precipitation. We examine a multivariate time series with feature vectors $(\mathbf{x}_1, \ldots, \mathbf{x}_T)$, where each $\mathbf{x}_t \in \mathbb{R}^m$ contains the feature values at time step $t$, and $T$ represents the total length of the observed time series. The corresponding univariate target series is $(Y_1, \ldots, Y_T)$, where $Y_t$ denotes the precipitation value at time step $t$.

Our objective is to predict the future $k$ steps of the target series using $h$ steps of historical multivariate data. In particular, while the input features $\mathbf{x}_{t}$ capture a variety of meteorological variables, the target $Y_t$ represents precipitation, which is heavily zero-inflated. This setting introduces additional difficulty compared to standard forecasting tasks: the model must not only learn complex temporal dependencies across multiple variables, but also distinguish the rare non-zero precipitation events from the overwhelming majority of zeros.

\subsection{Why We Cannot Apply Transformer Directly}

When dealing with zero-inflated data, where the target variable contains a large proportion of zero values, Transformer-based models face significant challenges in learning effective attention patterns. Our analysis shows that the overwhelming dominance of zeros creates a biased attention distribution: the model disproportionately focuses on zero-valued time steps due to their numerical prevalence, while struggling to attend to the sparse but critical non-zero precipitation events. This limitation can be rigorously analyzed through the formulation of the self-attention mechanism.

In Transformer training, self-attention weights are computed as:
\begin{equation}
\text{Attention}(Q, K, V) = \text{softmax}\left(\frac{QK^T}{\sqrt{d_k}}\right)V,
\label{eq:attention}
\end{equation}
where $Q$, $K$, and $V$ denote the query, key, and value matrices, respectively, and $d_k$ is the dimension of the key vectors (not to be confused with the prediction horizon $k$). In the context of precipitation forecasting, $Q$ is derived from target precipitation values, $K$ from historical data, and $V$ from feature embeddings.

The attention distribution depends on the similarity score matrix $\frac{QK^T}{\sqrt{d_k}}$. 
In zero-inflated data, this similarity structure is distorted: embeddings corresponding to zero-valued targets tend to form a high-density cluster in the representation space, whereas embeddings for non-zero values are more dispersed. 
Formally, for any two time steps $i, j$, the similarity score is
\begin{equation}
    S_{ij} = \frac{Q_i \cdot K_j}{\sqrt{d_k}}.
\end{equation}
The attention weights are then defined as
\begin{equation}
    A_{ij} = \frac{\exp(S_{ij})}{\sum_{\ell}\exp(S_{i\ell})},
\end{equation}
where the denominator sums over all candidate keys $\ell$.

When both $i$ and $j$ correspond to zero-valued targets, their similarity scores concentrate around a constant with only small variance:
\begin{equation}
S_{ij}|_{Y_i=0, Y_j=0} \approx c_0 \pm \epsilon_0,
\end{equation}
where $\epsilon_0$ is small. In contrast, when at least one of the time steps corresponds to a non-zero target, the variance in similarity scores is much larger. Although higher variance in principle reflects richer information for distinguishing meaningful events, because zeros vastly outnumber non-zeros, the softmax normalizer is dominated by zero positions; this offsets the potential benefit of higher variance, and the total attention on non-zero positions still shrinks as $\pi_0$ increases.

Passing through the softmax function amplifies this effect. For a fixed query $i$, the total attention mass assigned to non-zero events is
\begin{equation}
\sum_{j:Y_j>0} A_{ij}
= \frac{\sum_{j:Y_j>0}\exp(S_{ij})}{\sum_{j:Y_j=0}\exp(S_{ij}) + \sum_{j:Y_j>0}\exp(S_{ij})},
\end{equation}
which \textbf{decreases monotonically with the zero-inflation ratio $\pi_0$}. This means that the higher the zero ratio, the less attention mass non-zero events receive, leading to systematic under-attention to the minority but crucial events.

Beyond the direct reduction of attention mass on non-zero events, 
the global Shannon entropy provides a complementary perspective: 
it quantifies how the attention distribution collapses toward zero positions as $\pi_0$ increases. 
Formally, the entropy of the attention distribution is 
$H(A_{i\cdot}) = -\sum_j A_{ij}\log A_{ij}$. 
Its behavior with respect to $\pi_0$ is not strictly monotonic: 
within the zero block, similarity scores are nearly equal, inducing a near-uniform allocation that can raise entropy, 
whereas globally the dominance of zeros in the softmax normalizer skews the allocation toward zero positions, lowering entropy.

This analysis is consistent with recent findings in the literature. For example, in \cite{liu2023nonstationarytransformersexploringstationarity}, it is shown that with the \textit{Non-stationary Transformer} shows that attention can fail under non-stationary time series because the similarity matrix drives an \textit{attention distribution concentrated on short-horizon neighbors}, impairing cross-scale aggregation of temporal dependencies. Although the underlying cause differs (non-stationarity versus zero inflation), both cases reveal a common mechanism: attention collapses toward dominant but less informative structures. Similarly, \cite{zhai2023stabilizing} identifies the phenomenon of \textit{attention entropy collapse}, where over-concentration of attention leads to instability during training and reduced generalization ability. These results corroborate our observation that attention in zero-inflated precipitation forecasting tends to concentrate on less-informative structures, reflecting a structural bias in self-attention.

These findings highlight the inherent difficulty of applying standard self-attention mechanisms to zero-inflated data, motivating the development of our proposed framework to mitigate these challenges. Additional analyses of other advanced Transformer-based models, including the Non-stationary Transformer, are provided in the github\footnote{https://github.com/Wentao-Gao/ZIDF-from-noise-to-precision} for a comprehensive understanding.

\section{Zero inflation diffusion framework}

\begin{figure*}[h]
    \centering
    \includegraphics[width=0.9\linewidth]{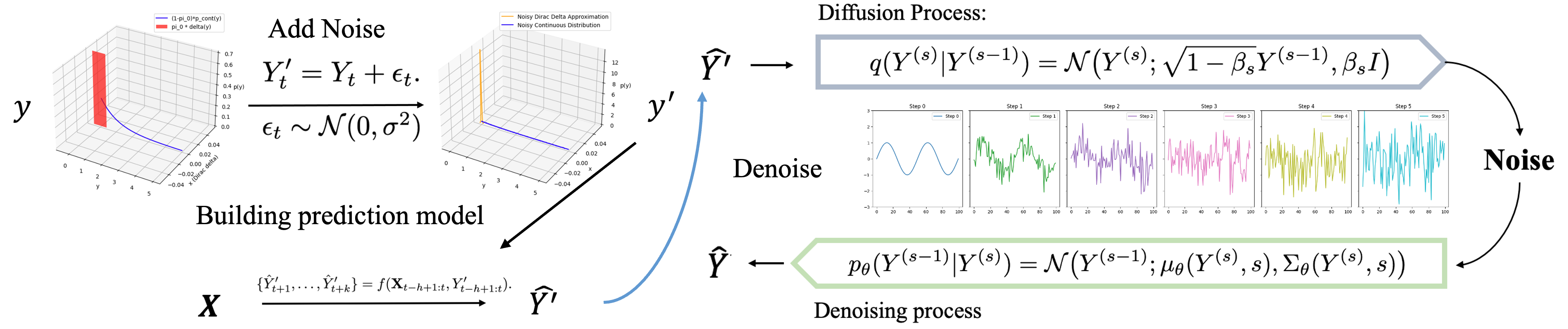}
     \caption{The overall architecture of the Zero-Inflated Diffusion Framework (ZIDF) for modeling sparse precipitation events. The process starts by adding Gaussian noise to the zero-inflated target \(Y\) to generate a smoothed distribution \(Y'\). Using historical features \(\mathbf{X}\), the prediction model forecasts noisy future values \(\hat{Y}'\). The diffusion process models the forward noise distribution, while the denoising process recovers the predicted target \(\hat{Y}\).}
    \label{fig:process}
\end{figure*}

This section firstly introduces the proposed ZIDF (Zero inflation diffusion framework) for addressing zero inflation in time series data. As shown in Figure \ref{fig:process}, ZIDF comprises three main components: noise injection, predictive modeling, and diffusion-based denoising. With noise injection, we transform the non-differentiable zero-inflated distribution into a smooth continuous distribution via Gaussian convolution. We then train a predictive component that utilizes both input features and past target information on these smoothed data, allowing stable gradient-based optimization. Finally, we employ a diffusion-based denoising procedure to recover the original distribution from the model's noisy predictions. In the following, we present the components in detail (Sections 3.1 to 3.3). Then in Section 3.4, we provide a theoretical guarantee showing that the denoised predictions coincide with those one would obtain using the clean, non-inflated targets, under suitable assumptions. The implementation details of ZIDF are provided in Section 3.5.

\subsection{From Non-Differentiable to Differentiable Distributions}

Zero-inflated time series data often exhibit a point mass at zero. In the context of precipitation forecasting, where precipitation exhibits zero inflation, the target variable \( Y \) can be modeled as:
\begin{equation}
p_{\text{data}}(y) = \pi_0 \delta(y) + (1-\pi_0)p_{\text{cont}}(y),
\end{equation}
where \(\pi_0 = P(Y=0)\) is the zero inflation ratio, \(\delta(\cdot)\) is the Dirac delta function, and \(p_{\text{cont}}(y)\) is a continuous density for \(y > 0\). This formulation is widely adopted in the literature for modeling zero-inflated data, as it effectively captures the prevalence of zero observations alongside a continuous distribution for positive values \cite{lambert1992zero,rose2006use}.

The Dirac delta function \(\delta(y)\) introduces a non-differentiable component at zero, making gradient-based optimization difficult. To alleviate this issue, we add Gaussian noise \(\epsilon \sim \mathcal{N}(0,\sigma^2)\):
\begin{equation}
Y' = Y + \epsilon.
\end{equation}
This yields a convolved distribution:
\begin{equation}
q_{\sigma}(y') = \int_{-\infty}^{\infty} p_{\text{data}}(y) g_{\sigma}(y'-y) \, dy,
\end{equation}
where \( q_{\sigma}(y) \) represents the distribution of the perturbed target variable \( Y' \) and \(g_{\sigma}\) is a Gaussian kernel. Gaussian convolution smooths the point mass, ensuring \(q_{\sigma}(y')\) is infinitely differentiable, thus suitable for stable gradient-based training.

\begin{lemma}
After Gaussian convolution, the resulting distribution $q_{\sigma}(y')$ is infinitely differentiable \cite{rudin1987real,stein2003fourier,folland1999real}:
\begin{equation}
q_{\sigma}(y') = \pi_0 g_{\sigma}(y') + (1-\pi_0)(p_{\text{cont}} * g_{\sigma})(y').
\end{equation}

Intuitively, the Gaussian kernel $g_{\sigma}$ acts as a smoothing operator that transforms the discontinuous Dirac delta function into a smooth Gaussian bump. This smoothness property is crucial for gradient-based optimization, as it ensures well-defined gradients throughout the parameter space. (Full proof in github section 1 of the Appendix).
\end{lemma}

\subsection{Training on Smoothed Targets}
With the smoothed, differentiable targets \(\{Y_t'\}_{t=1}^{T}\), we train the predictive component \( f \) of ZIDF framework. This component takes the historical features \(\mathbf{X}_{t-h+1:t}\) and the corresponding smoothed target values \( Y_{t-h+1:t}' \) as input to generate intermediate noisy predictions:
\begin{equation}
\hat{Y}_{t+1:t+k}' = f(\mathbf{X}_{t-h+1:t}, Y_{t-h+1:t}').
\end{equation}

To train the predictive component, we minimize the mean squared error (MSE) between the smoothed targets and the intermediate noisy predictions:
\begin{equation}
\mathcal{L}_{\text{pred}} = \frac{1}{T-h-k+1} \sum_{t=h}^{T-k} \sum_{j=1}^k \big(Y_{t+j}' - \hat{Y}_{t+j}'\big)^2.
\end{equation}

This process facilitates stable gradient-based optimization, enabling the predictive component to effectively capture meaningful temporal patterns despite the challenges of zero-inflated data.

\subsection{Diffusion-Based Denoising}

The diffusion-based denoising process in ZIDF framework then refines these noisy predictions \(\hat{Y}_{t+1:t+k}'\) to recover a distribution that aligns with the original target \(\hat{Y}_{t+1:t+k}\).

After training on smoothed data, the model produces noisy predictions \(\hat{Y}_{t+1:t+k}'\). To recover the clean target distribution from these noisy predictions, we employ a Denoising Diffusion Probabilistic Model (DDPM) \cite{ho2020denoising}.

The core idea of DDPMs is to define a forward diffusion process that gradually maps arbitrary data distributions into a standard Gaussian distribution, and a reverse diffusion process that learns to reconstruct the original data distribution. Below, we describe how this applies to our noisy predictions \(\hat{Y}_{t+1:t+k}'\) .

\paragraph{Forward Diffusion Process}
Starting from a clean data sample \(Y^{(0)}\) drawn from the data distribution \(p_{\text{data}}(y)\), the forward diffusion process introduces noise step by step:
\begin{equation}
q(Y^{(s)} | Y^{(s-1)}) = \mathcal{N}\bigl(Y^{(s)}; \sqrt{1 - \beta_{s}} Y^{(s-1)}, \beta_{s} I \bigr),
\end{equation}
where \(\{\beta_{s}\}_{s=1}^S\) is a predefined noise schedule. Define \(\alpha_{s} = 1 - \beta_{s}\) and \(\bar{\alpha}_{s} = \prod_{i=1}^{s} \alpha_i\), then:
\begin{equation}
q(Y^{(s)} | Y^{(0)}) = \mathcal{N}\bigl(Y^{(s)}; \sqrt{\bar{\alpha}_{s}} Y^{(0)}, (1 - \bar{\alpha}_{s}) I \bigr).
\end{equation}
As \(s \to S\), \(Y^{(S)}\) converges to a standard Gaussian distribution \(\mathcal{N}(0, I)\).

Note that the diffusion model is trained on the original clean precipitation data \(Y\), not on the already noisy data \(Y'\). This separation allows the diffusion model to learn the natural data distribution and effectively denoise the predictions from the predictive component.

\paragraph{Reverse Diffusion Process}
The reverse process approximates the true reverse conditional distributions:
\begin{equation}
p_\theta(Y^{(s-1)} | Y^{(s)}) \approx q(Y^{(s-1)} | Y^{(s)}, Y^{(0)}).
\end{equation}
Assuming a Gaussian form:
\begin{equation}
p_\theta(Y^{(s-1)} | Y^{(s)}) = \mathcal{N}\bigl(Y^{(s-1)}; \mu_\theta(Y^{(s)}, s), \Sigma_\theta(Y^{(s)}, s)\bigr),
\end{equation}
where \(\Sigma_\theta\) is typically fixed. Instead of directly predicting \(\mu_\theta\), the model learns to predict the noise \(\epsilon\) added during forward diffusion:
\begin{equation}
Y^{(s)} = \sqrt{\bar{\alpha}_{s}} Y^{(0)} + \sqrt{1 - \bar{\alpha}_{s}} \epsilon, \quad \epsilon \sim \mathcal{N}(0, I).
\end{equation}
The model estimates \(\epsilon_\theta(Y^{(s)}, s)\), and the mean \(\mu_\theta\) can be written as:
\begin{equation}
\mu_\theta(Y^{(s)}, s) = \frac{1}{\sqrt{\alpha_{s}}} \left(Y^{(s)} - \frac{1 - \alpha_{s}}{\sqrt{1 - \bar{\alpha}_{s}}} \epsilon_\theta(Y^{(s)}, s) \right).
\end{equation}

\paragraph{Training Objective}
The training objective for DDPM minimizes the variational lower bound (VLB) on the negative log-likelihood, which simplifies to:
\begin{equation}
\mathcal{L}_\text{DDPM} = \mathbb{E}_{s, Y^{(0)}, \epsilon} \left[ \left\| \epsilon - \epsilon_\theta\left(\sqrt{\bar{\alpha}_{s}} Y^{(0)} + \sqrt{1 - \bar{\alpha}_{s}} \epsilon, s \right) \right\|^2 \right].
\end{equation}

\paragraph{Denoising the Predictions \(\hat{Y}_{t+1:t+k}'\)}
At inference time, we treat the noisy predictions \(\hat{Y}_{t+1:t+k}'\) from the predictive model as the high-noise state \(Y_{t+1:t+k}^{(S)}\) in the diffusion framework. We then run the learned \textit{reverse diffusion} process to obtain the denoised predictions:
\begin{equation}
    Y_{t+1:t+k}^{(0)} = \text{ReverseDiffusion}(Y_{t+1:t+k}^{(S)}),
\end{equation}
where \(Y_{t+1:t+k}^{(0)}\) corresponds to the final cleaned output \(\hat{Y}_{t+1:t+k}\).

The noise level in the predictions \(\hat{Y}_{t+1:t+k}'\) naturally matches the expected noise level of the diffusion model's high-noise state, as both are derived from the same Gaussian perturbation process with noise parameter \(\alpha_{\text{noise}}\).

By design, the diffusion-based denoising procedure mitigates the noise introduced during prediction, ensuring that \(\hat{Y}_{t+1:t+k}\) aligns with the clean target distribution. Formally,
\[
\mathbb{E}\bigl[\| \hat{Y}_{t+1:t+k}' - \hat{Y}_{t+1:t+k}\|^2\bigr] \;\to\; 0,
\]
indicating that the denoised predictions converge in expectation to their noise-free counterparts.

For precipitation forecasting applications, an additional consideration is ensuring physically meaningful (non-negative) outputs. The diffusion model is trained on non-negative precipitation data, which encourages realistic value generation. To guarantee strict non-negativity, we apply a simple $\max(0, \cdot)$ operation to the final denoised predictions.

\subsection{Theoretical Guarantee for Denoised Predictions}

We now formally present the effect of our smoothing and denoising framework as a theorem (Theorem 1), to provide the theoretical guarantee for the proposed approach. Firstly, we make the following assumption.

\begin{assumption}[Unbiasedness]
The predictive component $f$, trained on smoothed targets $Y'$, is unbiased in the sense that it learns the conditional mean of $Y'$ given the historical information. Formally,
\begin{equation}
\mathbb{E}[Y'_{t+1:t+k} \mid \mathcal H_t] 
= f(Y'_{t-h+1:t}, \mathbf{X}_{t-h+1:t}),
\end{equation}
where $\mathcal H_t = \sigma(Y'_{t-h+1:t}, \mathbf{X}_{t-h+1:t})$ denotes the $\sigma$-algebra generated by the (smoothed) historical targets and covariates.
\end{assumption}

\begin{theorem}
Under Assumption~1 and the denoising capability of the diffusion model, the final predictions after reverse diffusion satisfy:
\begin{equation}
\mathbb{E}[\hat{Y}_{t+1:t+k} \mid \mathcal H_t] 
= \mathbb{E}[Y_{t+1:t+k} \mid \mathcal H_t].
\end{equation}
That is, the denoised predictions $\hat{Y}$ are, in expectation, equal to the conditional mean of the clean targets $Y$ given the same historical information.
\end{theorem}

\begin{proof}[Proof Sketch]
By Assumption~1, Since $\hat Y'_{t+1:t+k}$ depends on $(Y'_{t-h+1:t},\mathbf X_{t-h+1:t})$, it is $\mathcal H_t$–measurable. To align this with the diffusion framework, we embed the prediction into the forward diffusion at step $\tau$ by setting \(Y^{(\tau)}=\sqrt{\bar\alpha_\tau}\,Y'_{t+1:t+k}+\sqrt{1-\bar\alpha_\tau}\,\xi,
\) where $\xi\sim\mathcal N(0,I)$ is an independent Gaussian noise, and extend the information set to $\bar{\mathcal H}_t=\sigma(\mathcal H_t,\xi)$. In this way, the model prediction is embedded into a specific noise stage of the diffusion process (the $\tau$-th step), ensuring consistency with the forward diffusion and allowing the reverse denoiser to be applied. From the standard DDPM identity under the simple loss, the reverse denoiser then outputs 
\(\hat Y_{t+1:t+k}=\mathbb E[\,Y_{t+1:t+k}\mid Y^{(\tau)},\bar{\mathcal H}_t\,],
\)
which represents the posterior mean of the clean target given the noisy input. Taking conditional expectations and applying the tower property yields $\mathbb E[\hat Y_{t+1:t+k}\mid \bar{\mathcal H}_t]=\mathbb E[Y_{t+1:t+k}\mid \bar{\mathcal H}_t]$. Finally, since the added noise $\xi$ is independent of the history, conditioning on $\bar{\mathcal H}_t$ is equivalent to conditioning on $\mathcal H_t$, giving $\mathbb E[\hat Y_{t+1:t+k}\mid \mathcal H_t]=\mathbb E[Y_{t+1:t+k}\mid \mathcal H_t]$. This shows that the denoised predictions match, in expectation, the conditional mean of the clean targets. A full proof is provided in the appendix.
\end{proof}

\subsection{Implementation Details}

\begin{figure}[h]
    \centering
    \includegraphics[width=0.9\linewidth]{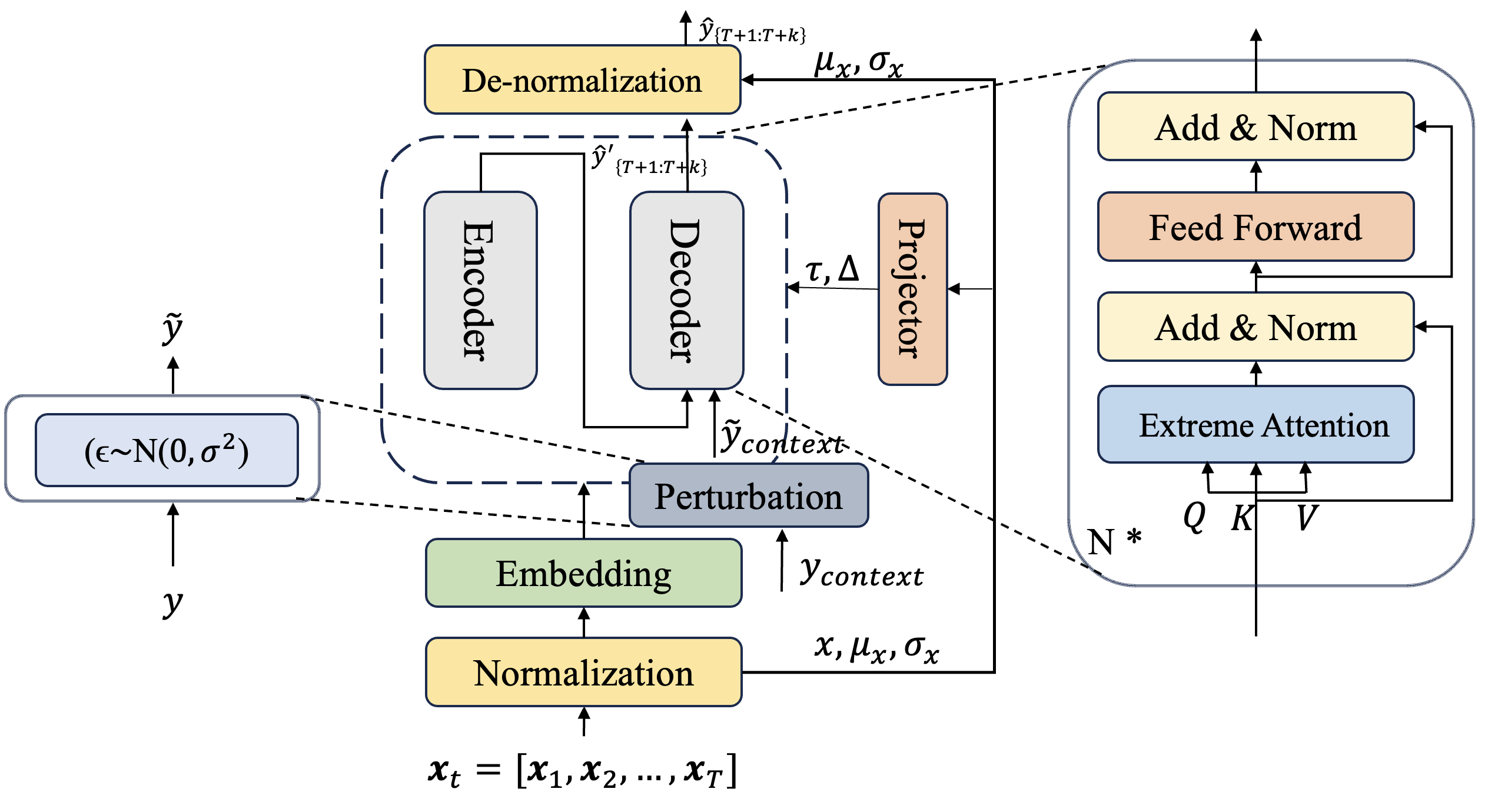}
    \caption{Architecture of Non-stationary Transformer \cite{liu2023nonstationarytransformersexploringstationarity}.}
    \label{fig:nonstation}
\end{figure}

\paragraph{Predictive Component}  
For the predictive component \( f \), we employ a Non-stationary Transformer architecture, as illustrated in Figure~\ref{fig:nonstation}. This model is designed to effectively capture temporal dependencies and address nonstationarities in real-world time series data. By leveraging historical targets \( Y_{t-k+1:t} \) and features \( \mathbf{X}_{t-k+1:t} \), the model achieves robust performance in predicting future intervals.

During testing, we apply the same Gaussian noise with parameter \(\alpha_{\text{noise}}\) to the historical target values to maintain consistency with the training condition. This ensures that the predictive component operates on data with the same statistical properties as seen during training.

The encoder and decoder of the Non-stationary Transformer consist of 2 and 1 layers, respectively. Following the default settings from \cite{liu2023nonstationarytransformersexploringstationarity}, the model has a hidden dimension of 512, 8 attention heads and a feedforward network dimension of 2048. The moving average window size is set to 25, with an attention factor of 1. GELU activation is used, and the encoder incorporates distillation (\texttt{distil=True}). Temporal features are encoded using the \texttt{timeF} method, and a dropout rate of 0.05 is applied to mitigate overfitting. Further details are provided in the github.

\begin{figure*}[h]
            \centering
            \includegraphics[width=0.9\linewidth]{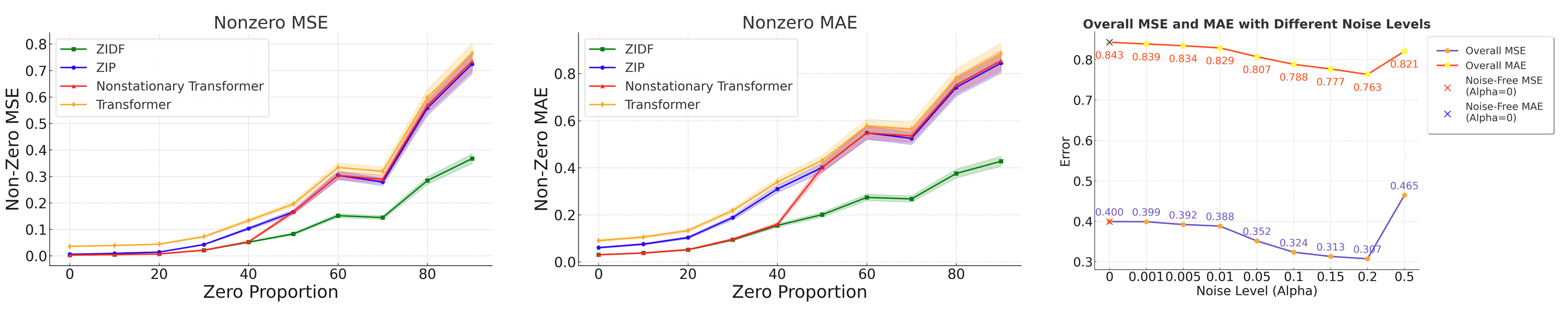}
            \caption{The figure illustrates the performance of different models under varying zero inflation proportions and noise levels. The left two panels display non-zero MAE and non-zero MSE across different zero proportions, showing that the ZIDF model consistently outperforms others, especially under higher zero proportions. The right panel analyzes overall MSE and MAE as a function of noise levels (alpha). Notably, the performance is optimal within the noise level range of 0.1 to 0.2, with minimal variation across different noise levels. This demonstrates the ZIDF model's low sensitivity to the noise parameter, highlighting its robustness. The noise-free scenario (alpha = 0) is marked for baseline error comparison.}
            \label{fig:simulation}
\end{figure*}

\paragraph{Denoising Model}  
We adopt a U-Net-based DDPM architecture for the denoising process \cite{ho2020denoising}, as outlined in Algorithm~1 in the Appendix. This architecture is chosen for its ability to perform multi-scale feature extraction. The U-Net encoder captures high-level temporal features by progressively aggregating information across different time scales, while the decoder reconstructs fine-grained temporal patterns with skip connections that retain detailed temporal dependencies. Positional embeddings encode the temporal order of the data.

The training process for ZIDF consists of two independent stages: (1) training the predictive component on noisy targets with noise level \(\alpha_{\text{noise}}\), and (2) training the diffusion model on clean historical data. During inference, we apply the same noise level \(\alpha_{\text{noise}}\) to historical observations before feeding them to the predictive component, ensuring consistency between training and testing conditions.

In the diffusion process, let \( s \) denote the diffusion step, ranging from \( s = 1 \) to \( s = S \), where \( S \) is the total number of diffusion steps. During training, noise \( \epsilon \sim \mathcal{N}(0, I) \) is sampled and added to a clean target \( Y^{(0)} \) to form \( Y^{(s)} \) using the forward diffusion process. The U-Net predicts the added noise \( \hat{\epsilon}(Y^{(s)}, s) \) for each \( s \), and the loss is computed as \( (\epsilon - \hat{\epsilon})^2 \), enabling the model to iteratively learn to denoise at each diffusion step.

During inference, starting from \( Y^{(S)} \sim \mathcal{N}(0, I) \), the learned reverse diffusion process is applied step-by-step to recover \( Y^{(0)} \), which aligns with the clean target distribution.

In DDPM, we use \( S = 1000 \), a linear noise schedule \( \beta_s \), a learning rate of \( 1 \times 10^{-4} \), and a batch size of 64. Training stability and efficiency are further enhanced through gradient clipping, cosine learning rate annealing, and data augmentation.

After generating noisy predictions \( \hat{Y}_{t+1:t+k}' \) from the predictive component \( f(Y_{t-h+1:t}, \mathbf{X}_{t-h+1:t}) \), the reverse diffusion process is applied to denoise the predictions and recover the original distribution.

\section{Experiments}
To evaluate the performance of our proposed method in solving the zero-inflation problem, we conduct both controlled simulation studies and a real-world case study. The simulations vary the proportion of zeros and the injected noise level to assess robustness, while the case study uses South Australian precipitation data and compares ZIDF with representative baselines (e.g., Non-stationary Transformer, iTransformer, ZIP/Hurdle) under MSE/MAE across multiple prediction horizons.

\subsection{Simulation Analysis}
\subsubsection{Impact of Zero Proportion on Model Performance}
    
The zero proportion variation experiment was conducted to evaluate the robustness and performance of the proposed method under a wide range of zero inflation levels in the target variable. Synthetic datasets with controlled zero proportions (\(\pi_0 \in \{0.0, 0.1, 0.2, \dots, 0.9\}\)) were generated to simulate multivariate time series prediction scenarios. The target variable, sampled from a Gamma distribution, represents precipitation intensity, with increasing sparsity systematically introduced as the zero proportion rises. Covariates were designed to capture realistic temporal dynamics, incorporating periodic, linear, and random trends, while maintaining meaningful relationships with the target variable.

The prediction task requires forecasting the next three timesteps of the zero-inflated target variable based on 12 steps historical observations. To benchmark the proposed method, it was compared against three representative baseline models: the standard Transformer \cite{vaswani2017attention}, the Non-stationary Transformer \cite{liu2023nonstationarytransformersexploringstationarity}, and the Zero-Inflated Poisson model \cite{lambert1992zero}. Model performance was evaluated using mean squared error (MSE) and mean absolute error (MAE) as metrics to comprehensively assess accuracy across varying zero proportions.

Results as shown in Figure~\ref{fig:simulation} demonstrate that the proposed method maintains stable performance under high zero inflation, consistently outperforming baseline models in MSE, MAE. These findings validate the adaptability of the method to high zero inflation scenarios and underscore its potential for real-world precipitation prediction tasks.

\subsubsection{Evaluating the Trade-off in Noise Injection for Model Performance}

This experiment evaluates the trade-off between noise injection and the recovery capability of the denoising process in handling zero-inflated time series data. A simulated dataset was generated, featuring a target variable with a 70\% zero proportion and non-zero values around \(10^{-1}\). Realistic covariates with periodic and linear trends were included to mimic real-world scenarios. Gaussian noise was injected into the target variable, with the noise intensity determined by:
\begin{equation}
    \sigma = \alpha_{\text{noise}} \cdot \mu_Y,
\end{equation}
where \(\alpha_{\text{noise}}\) is the noise ratio controlling the magnitude of the noise, and \(\mu_Y\) is the mean of the non-zero values in the target variable \(Y\). We tested multiple values of \(\alpha_{\text{noise}}\) within the range \([0.001, 0.005, 0.01, 0.05, 0.1, 0.15, 0.2, 0.5]\), ensuring the noise scale remained proportional to the inherent variability of the data. The noisy target variable \(Y'(t)\) was then calculated as:
\begin{equation}
Y'(t) = Y(t) + \epsilon(t),
\end{equation}
where \(\epsilon(t) \sim \mathcal{N}(0, \sigma^2)\). A diffusion model was applied to denoise the data and recover the original distribution.

The results, as shown in Figure~\ref{fig:simulation}, demonstrate that the model is robust to variations in the noise injection parameter \(\alpha_{\text{noise}}\). For \(\alpha_{\text{noise}}\) values between 0.1 and 0.2, both MSE and MAE remain near their minimum levels, indicating that the model's performance is not highly sensitive to the precise value of \(\alpha_{\text{noise}}\) within this range. This robustness simplifies the parameter tuning process, making the method practical for real-world applications where determining an ideal noise level can be challenging. By achieving stable performance across a range of \(\alpha_{\text{noise}}\) values, the proposed approach validates the effectiveness of the noise injection and denoising mechanism in preserving the target distribution while mitigating distortions.For simplicity, the noise injection parameter \(\alpha_{\text{noise}}\) is set to 0.1 in our real-world experiments. This value is used consistently during both training of the predictive component and inference, ensuring that the noise level in \(\hat{Y}'\) matches the expected input for the diffusion denoising process.

\begin{table*}[h!]
\centering
\caption{Long-term Forecast Results (Including ZIP and Hurdle Models)}
\label{tab:main_results}
\resizebox{\textwidth}{!}{%
\begin{tabular}{@{}llccccccccccc@{}}
\toprule
\textbf{Output Length} & \textbf{} & \textbf{ZIDF (Ours)} & \textbf{TSMixer} & \textbf{iTransformer} & \textbf{Transformer} & \textbf{PatchTST} & \textbf{TimeMixer} & \textbf{TimesNet} & \textbf{Non-stationary Transformer} & \textbf{ZIP} & \textbf{Hurdle} & \textbf{ZIG} \\ 
                       &           & \textbf{}                     & \textbf{(2022)}  & \textbf{(2023)}        & \textbf{(2017)}      & \textbf{(2023)}   & \textbf{(2022)}   & \textbf{(2023)}   & \textbf{(2023)}                   & \textbf{(1996)} & \textbf{(1998)} & \textbf{(2018)} \\ \midrule
\multirow{2}{*}{\textbf{24}} & \textbf{MSE} & \textbf{0.3785}          & 0.9458           & 0.9018                 & 0.8723               & 0.9065            & 0.9089            & 0.6520            & 0.8723                             & 1.5620       & 1.8725          & 1.6845          \\
                             & \textbf{MAE} & \textbf{0.2487}          & 0.4010           & 0.3256                 & 0.3157               & 0.3251            & 0.3140            & 0.4291            & 0.2975                             & 0.5612       & 0.6250          & 0.5835          \\ \midrule
\multirow{2}{*}{\textbf{48}} & \textbf{MSE} & \textbf{0.3842}          & 0.9491           & 0.9082                 & 0.8641               & 0.9118            & 0.9127            & 0.6620            & 0.8881                             & 1.6031       & 1.8952          & 1.7250          \\
                             & \textbf{MAE} & \textbf{0.2460}          & 0.4084           & 0.3171                 & 0.3312               & 0.3243            & 0.3211            & 0.4337            & 0.3117                             & 0.5763       & 0.6425          & 0.5980          \\ \midrule
\multirow{2}{*}{\textbf{96}} & \textbf{MSE} & \textbf{0.3956}          & 0.9546           & 0.9220                 & 0.8716               & 0.9185            & 0.9205            & 0.6776            & 0.9000                             & 1.7025       & 1.9331          & 1.8120          \\
                             & \textbf{MAE} & \textbf{0.2593}          & 0.4098           & 0.3299                 & 0.2973               & 0.3067            & 0.3156            & 0.4455            & 0.3078                             & 0.5889       & 0.6520          & 0.6125          \\ \midrule
\multirow{2}{*}{\textbf{192}} & \textbf{MSE} & \textbf{0.4060}         & 0.9671           & 0.9292                 & 0.8764               & 0.9200            & 0.9252            & 0.7450            & 0.8900                             & 1.7501       & 1.9803          & 1.8510          \\
                              & \textbf{MAE} & \textbf{0.2640}         & 0.4053           & 0.3339                 & 0.3090               & 0.3200            & 0.3220            & 0.4904            & 0.2984                             & 0.6025       & 0.6624          & 0.6275          \\ \midrule
\multirow{2}{*}{\textbf{336}} & \textbf{MSE} & \textbf{0.4130}         & 0.9734           & 0.9215                 & 0.8777               & 0.9250            & 0.9267            & 0.7500            & 0.8973                             & 1.8457       & 2.0310          & 1.9265          \\
                              & \textbf{MAE} & \textbf{0.2684}         & 0.4253           & 0.3217                 & 0.3048               & 0.3150            & 0.3235            & 0.4950            & 0.3026                             & 0.6205       & 0.6742          & 0.6450          \\ \midrule
\multirow{2}{*}{\textbf{720}} & \textbf{MSE} & \textbf{0.4212}         & 0.9733           & 0.9179                 & 0.8655               & 0.9300            & 0.9493            & 0.7600            & 0.9050                             & 1.9123       & 2.1057          & 1.9856          \\
                              & \textbf{MAE} & \textbf{0.2720}         & 0.4282           & 0.3150                 & 0.3050               & 0.3200            & 0.4196            & 0.5000            & 0.3080                             & 0.6358       & 0.6820          & 0.6592          \\ \bottomrule
\end{tabular}%
}
\end{table*}

\subsection{Real World Case Study}

\subsubsection{Data Preprocessing}

Our study focused on South Australia, a region characterized by significant variability in precipitation patterns and frequent drought conditions. South Australia's climate is predominantly semi-arid to arid, with notable precipitation variability influenced by seasonal weather systems such as the southern jet stream and tropical moisture influx, making it an ideal testbed for evaluating zero-inflated forecasting methods.

\begin{figure}[h]
    \centering
    \includegraphics[width=0.8\linewidth]{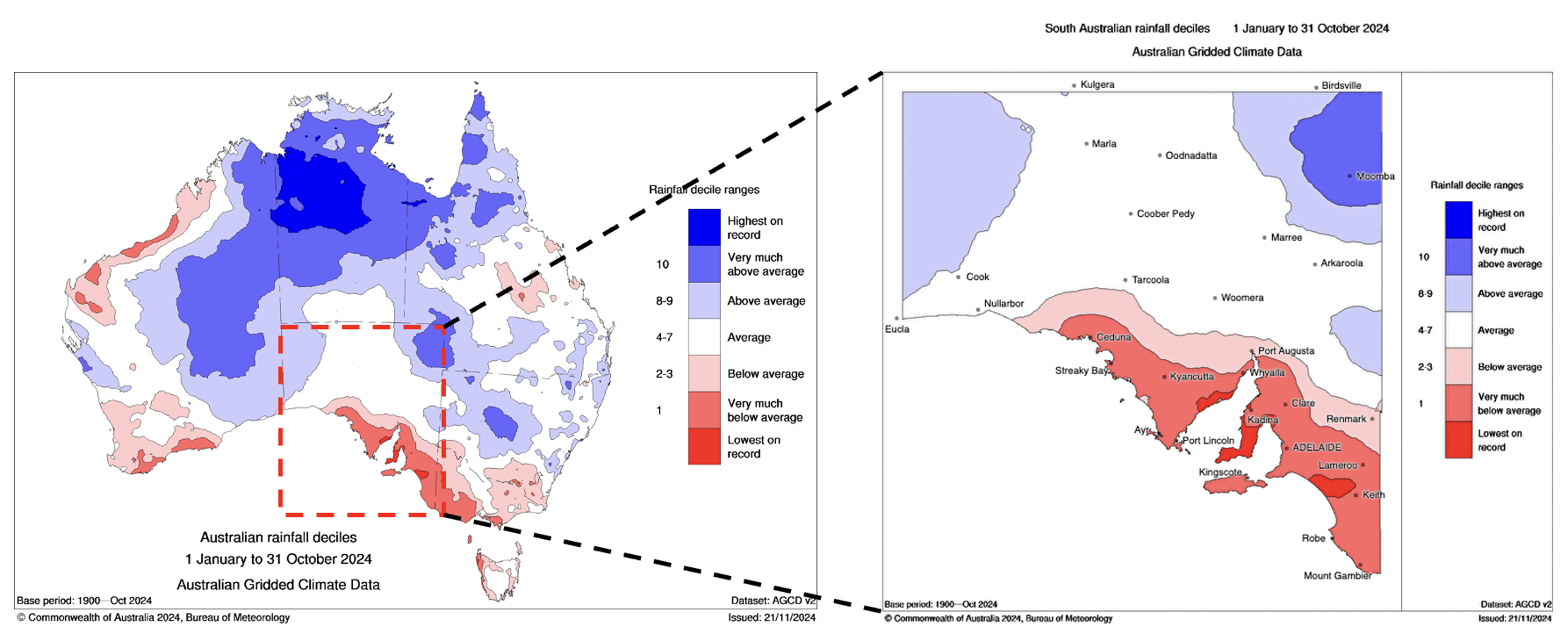}
    \caption{Study Area: South Australia (34°S–38°S, 129°E–141°E). The state is characterized by semi-arid to arid climate conditions with highly variable precipitation patterns.}
    \label{fig:study_area}
\end{figure}

We extracted and converted climate data from 1948 to 2014 for the South Australia region using the NCEP/NCAR Reanalysis dataset\footnote{\url{https://www.psl.noaa.gov/data/gridded/data.ncep.reanalysis.html}} \cite{kalnay1996ncep}, as shown in Figure \ref{fig:study_area}. The South Australia precipitation dataset exhibited significant zero inflation, with 68.3\% of daily observations recording zero precipitation. The data showed strong seasonal patterns, with winter months (June-August) having the highest precipitation frequency (45\% non-zero days) and summer months (December-February) showing the most severe zero inflation (85\% zero days).

Our features included relevant meteorological variables such as temperature, pressure, and wind speed, among others. The data, originally in NetCDF format \cite{netcdf4}, were transformed into CSV files for further processing. Detailed data preprocessing procedures were provided in the Appendix.

\subsubsection{Baselines}
To validate the effect of our method, we carefully chose 7 well-acknowledged forecasting models and 3 zero inflated models as our benchmark, including TSMixer \cite{chen2023tsmixerallmlparchitecturetime}, iTransformer \cite{liu2023itransformer}, Transformer \cite{vaswani2017attention}, PatchTST \cite{nie2023timeseriesworth64}, TimeMixer \cite{wang2024timemixerdecomposablemultiscalemixing}, TimesNet \cite{wu2023timesnettemporal2dvariationmodeling}, Non-stationary Transformer \cite{liu2023nonstationarytransformersexploringstationarity}, Zero Inflated Poisson (ZIP) \cite{lambert1992zero}, Zero Inflated Gaussian (ZIG) \cite{Hegde2018VariationalZG}, and Hurdle Model \cite{rose2006use} to ensure a comprehensive evaluation of our approach.

\subsubsection{Main Results}

In our experiments, the input sequence length was set to 48 steps. This choice was informed by prior studies, which demonstrated that the input length did not significantly impact model performance \cite{nie2023timeseriesworth64}. Moreover, excessively long input sequences could substantially increase computational complexity and training time. Therefore, an input length of 48 provided a reasonable balance between efficiency and effectiveness.

The output sequence lengths were set to 24, 48, 96, 192, 336, and 720 steps (days) to comprehensively evaluate the performance of the proposed method across different prediction horizons. These output lengths covered a range from short-term to long-term forecasts, ensuring that the experimental results were broadly applicable and reliable. In practical applications, short-term predictions were typically used for weather forecasting and agricultural management, while long-term forecasts were more suited for water resource planning and climate studies.

The results demonstrated that, while error metrics such as MSE and MAE generally increased with longer prediction horizons, our model consistently outperformed baseline models across all output lengths. Notably, it exhibited significant advantages in longer sequences, such as 336 and 720 steps. This highlighted the effectiveness of our approach in capturing complex temporal dependencies and addressing the challenges of zero inflation. Table~\ref{tab:main_results} presents detailed performance comparison across all models and prediction horizons.

\subsection{Ablation Study}

To validate the effectiveness of each component in our ZIDF framework, we conducted an ablation study by systematically removing key components. We used the 96-step prediction setting from our main experiments for this analysis. 
Table~\ref{tab:ablation} presents the detailed results.

The ablation study demonstrated the importance of each component in ZIDF. Removing Gaussian noise injection resulted in the most significant performance degradation (72.7\% increase in MSE), highlighting its crucial role in handling zero-inflated distributions. The diffusion-based denoising process also contributed substantially, reducing MSE by 30.9\%. Using a standard Transformer instead of the Non-stationary Transformer led to a 14.2\% increase in MSE, while reducing diffusion steps from 1000 to 100 increased MSE by 10.9\%. These results confirmed that all components of ZIDF worked synergistically to achieve optimal performance for zero-inflated precipitation prediction.

\begin{table}[h]
\centering
\caption{Ablation Study Results (96-step prediction)}
\label{tab:ablation}
\begin{tabular}{lcc}
\toprule
\textbf{Model Variant} & \textbf{MSE} & \textbf{MAE} \\
\midrule
ZIDF (Full Model) & \textbf{0.3956} & \textbf{0.2593} \\
\midrule
w/o Gaussian noise injection & 0.6832 & 0.3541 \\
w/o diffusion denoising & 0.5724 & 0.3162 \\
w/ standard Transformer & 0.4521 & 0.2854 \\
w/ S=100 (instead of S=1000) & 0.4389 & 0.2854 \\
\midrule
Non-stationary Transformer (baseline) & 0.9000 & 0.3078 \\
\bottomrule
\end{tabular}
\end{table}

\section{Conclusion}

This study introduces \textbf{ZIDF} for handling high zero values in precipitation forecasting. Combining Gaussian noise injection, smoothed-target training, and diffusion-based denoising, ZIDF robustly predicts zero-inflated distributions in synthetic and real-world settings. While achieving notable gains, it faces longer inference time ($\approx 3\times$ standard Transformers), performance drop under extreme zeros ($>90\%$), and uncertain cross-domain generalizability. The adaptable framework fits various forecasting models and excels in long-term prediction. Future work will focus on improving efficiency, designing adaptive noise schemes, and validating across domains.



\begin{ack}
This work was supported by the ARC Discovery Project DP230101122 and the University of South Australia Research Training Program (RTP) Scholarship. We gratefully acknowledge the continued support from the CSIRO Environment Research Unit and the Data61 Business Unit.
\end{ack}



\bibliography{m6636}

\end{document}